%% file: main.tex
\documentclass[10pt,twocolumn,letterpaper]{article}

\usepackage{multirow}
\usepackage{pifont}
\usepackage{booktabs}
\usepackage{todonotes}
\usepackage{amsmath}
\usepackage{bm}
\usepackage{graphics}
\usepackage{makecell}
\usepackage{lipsum}
\usepackage{color, colortbl}
\usepackage{enumitem} 
\usepackage{booktabs}
\usepackage{multirow}
\usepackage{rotating}
\usepackage{adjustbox}
\def\eg{\emph{e.g}\onedot} 

\def\ie{\emph{i.e}\onedot} 

\usepackage{mathtools}
\usepackage{textcomp}
\usepackage{gensymb}
\usepackage{lmodern}
\usepackage[accsupp]{axessibility} 

\usepackage[pagenumbers]{cvpr} 


\definecolor{mylightgray}{gray}{0.92}

\definecolor{cvprblue}{rgb}{0.21,0.49,0.74}
 
\def\eg{\emph{e.g}\onedot}

\def\ie{\emph{i.e}\onedot}

\usepackage[pagebackref,breaklinks,colorlinks,citecolor=cvprblue]{hyperref}

\def\paperID{8} 
\def\confName{CVPR}
\def\confYear{2024}

\title{What does CLIP know about peeling a banana?}

\author{\hspace{2mm}Claudia Cuttano$^1$, Gabriele Rosi$^{1,2}$, Gabriele Trivigno$^1$, Giuseppe Averta$^{1,2}$ \\
$^1$ Politecnico di Torino, $^2$ Focoos AI\\
$^1$ {\tt\small name.surname@polito.it}, $^2$ {\tt\small name.surname@focoos.ai}
}

\begin{document}

\twocolumn[{%
\renewcommand\twocolumn[1][]{#1}%
\maketitle
\begin{center}
    \includegraphics[width=0.78\linewidth]{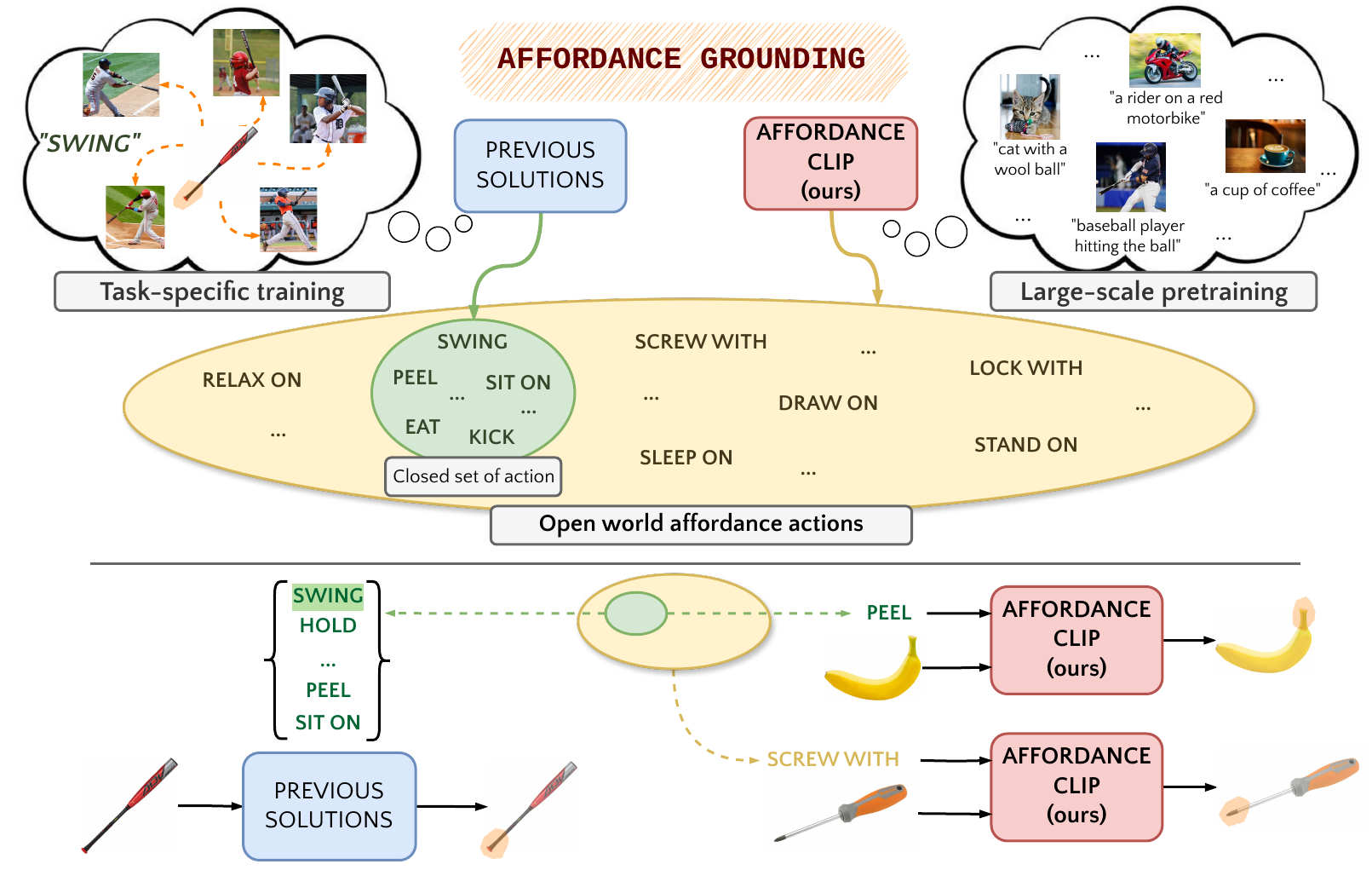}
    \captionof{figure}{\textbf{Overview of AffordanceCLIP}. Our AffordanceCLIP unlocks the hidden affordance understanding capabilities within CLIP. Traditional techniques rely on task-specific supervised training, limiting them to a closed set of actions. Our key insight is that CLIP, instead, already embeds knowledge on how humans interact with objects, without the need for explicit finetuning.  This enables open-vocabulary reasoning about a vast range of potential actions. Our open-vocabulary approach demonstrates promising performance in zero-shot, paving the way for broader and more flexible affordance understanding.
    }
    \label{fig:teaser}
    \vspace{0.5em}
\end{center}%
}]
\maketitle

\input{sec/0_abstract}    
\input{sec/1_intro}

\begin{figure*}[t]
    \centering
    \includegraphics[width=1\linewidth]{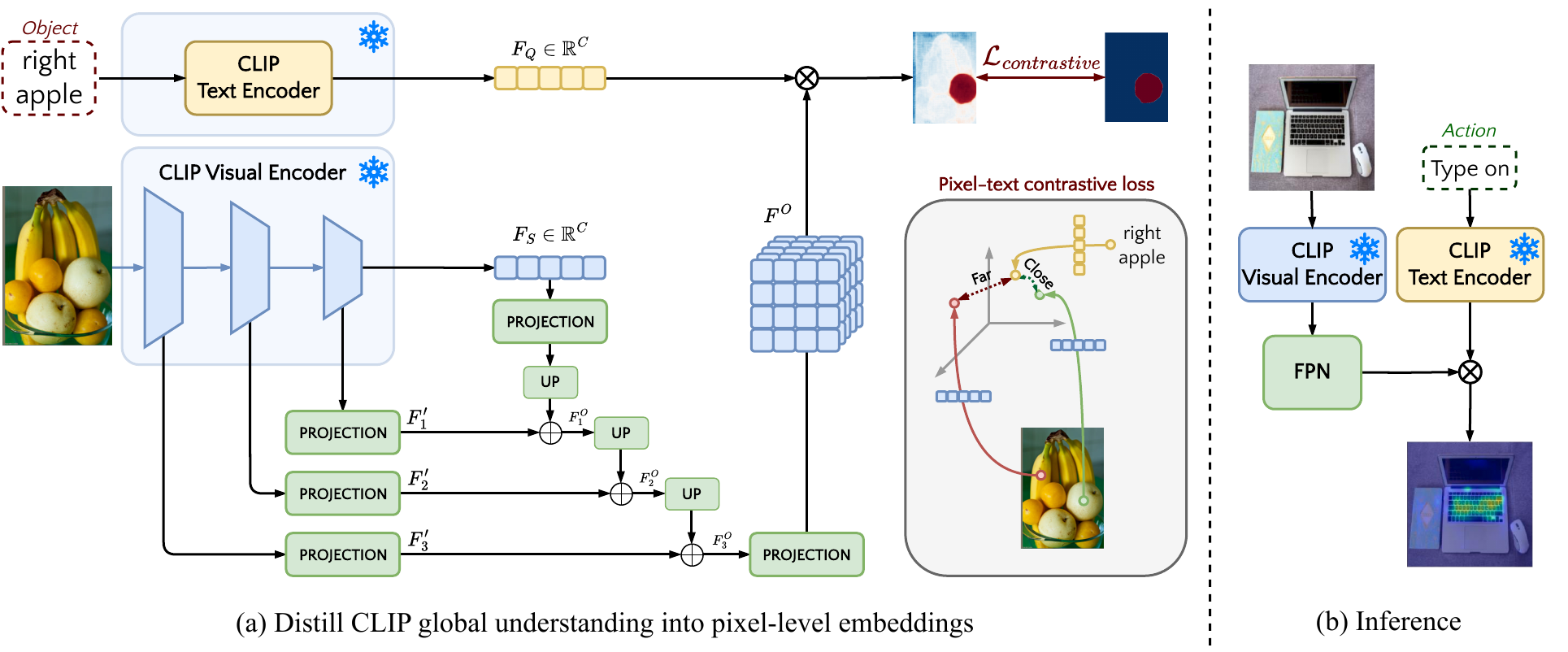}
    \caption{\textbf{Overview of the proposed AffordanceCLIP}. \textit{Left}: 
    We train a lightweight FPN to obtain dense feature maps from CLIP.  Given an image, and a textual query referring an object, a frozen CLIP model extracts visual and linguistic features. Then, our FPN gradually refines the output visual vector with fine-grained spatial details, in order to retain both spatial information and local image semantics. Finally, a contrastive loss encourages pixel-level embeddings within the  GT mask of the object to align with the corresponding linguistic features.     
    \textit{Right:}
    At inference, AffordanceCLIP can be directly queried with any textual prompt to obtain zero-shot affordance predictions.
    } 
    \label{fig:architecture}
\end{figure*}

\input{sec/2_related}
\input{sec/3_method}
\input{sec/4_experiments}
\input{sec/5_results}
\input{sec/6_ack}
{
    \small
    \bibliographystyle{ieeenat_fullname}
    \bibliography{main}
}


\end{document}

%% file: sec/0_abstract.tex
\begin{abstract}
\vspace{-7pt}
Humans show an innate capability to identify tools to support specific actions. The association between objects parts and the actions they facilitate is usually named \emph{affordance}. Being able to segment objects parts depending on the tasks they afford is crucial to enable intelligent robots to use objects of daily living. Traditional supervised learning methods for affordance segmentation require costly pixel-level annotations, while weakly supervised approaches, though less demanding, still rely on object-interaction examples and support a closed set of actions. These limitations hinder scalability, may introduce biases, and usually restrict models to a limited set of predefined actions. 
This paper proposes AffordanceCLIP, to overcome these limitations by leveraging the implicit affordance knowledge embedded within large pre-trained Vision-Language models like CLIP. We experimentally demonstrate that CLIP, although not explicitly trained for affordances detection, retains valuable information for the task.
Our AffordanceCLIP achieves competitive zero-shot performance compared to methods with specialized training, while offering several advantages: i) it works with any action prompt, not just a predefined set; ii) it requires training only a small number of additional parameters compared to existing solutions and iii) eliminates the need for direct supervision on action-object pairs, opening new perspectives for  functionality-based reasoning of models.

\end{abstract}

%% file: sec/1_intro.tex
\section{Introduction}
\label{sec:intro}

Our daily lives are filled with objects and tools that we effortlessly manipulate to achieve goals. Our natural ability to link  visual properties of an object (shape, material, and parts) with the actions it affords demonstrates a deep connection between perception and action. A concave shape, for instance, immediately suggests the ability to hold liquids, regardless of whether it is a cup or a coconut shell. 

In artificial intelligence, the problem of  
associating functionality with objects is known as affordance grounding \cite{ardon2020affordances, hassanin2021visual}. It aims at locating the regions of an object that can be used to carry out a given action. To date, standard approaches \cite{myers2015affordance,koppula2013learning, do2018affordancenet, chuang2018learning, qian2024affordancellm} attempt to solve this problem with supervised learning, relying on manually annotated datasets to teach models about object functionalities. Each object in the picture should be provided with multiple segmentation masks, one for each ``part'' of the object associated with functional tasks \cite{myers2015affordance, nguyen2017object}. In the case of a glass, we may imagine to have the edge associated to the action \emph{drink} and the handle with the action \emph{hold}. 
This paradigm, while effective, presents practical limitations due to the resource-intensive nature of acquiring pixel-level annotations.
Recognizing the need for more scalable and practical solutions, alternative techniques \cite{li2023locate,sawatzky2017weakly,nagarajan2019grounded,ag_from_exocentric_imgs,ag_from_exocentric_imgs+} formulated the affordance problem as a weakly supervised task, where the focus shifts towards learning object affordances through the observation of human-object interaction images \cite{burke2010neural}.  For example, a model might learn how to \textit{swing} a baseball bat after observing multiple images of humans grasping the bat (see \cref{fig:teaser}, top-left). 

Even though the annotations are simplified, we argue that weakly supervised approaches come with several limitations.  First, these approaches mainly work with images representing a single object (\eg a foreground baseball bat alone), limiting their use in real-world scenes with multiple objects. Additionally, they are trained on a closed-set of affordance actions (\ie 36 on the popular AGD20K \cite{ag_from_exocentric_imgs}), and cannot be used in an open vocabulary setting with arbitrary actions. Finally, in order to avoid introducing culture-dependent biases, they require a representative number of examples to learn from. For instance, the ways of carrying bags or chopping vegetables can be heavily influenced by the cultural habits \cite{plizzari2023can}.

In this paper we investigate whether it is possible to transfer affordance knowledge without direct supervision on a predefined set of classes. Our intuition relies on the observation that large pre-trained models may have already learnt how humans interact with functional objects by looking at millions of images. Even without datasets explicitly focused on affordances, these models potentially hold the key to identify affordances across a broader spectrum of actions than what can be achieved with annotated datasets.
To assess the validity of this hypothesis, we experiment with CLIP \cite{radford2021learning}, one of the most popular large Vision-Language model.
However, unlocking affordance knowledge from CLIP is non-trivial, as it aligns the image representations with textual descriptions on a global level, discarding spatial information. This makes its embeddings unsuitable for the affordance grounding task, which requires to localize specific object details depending on the textual prompt. 

Despite this, CLIP rich exposure to complex scenes and descriptive natural language suggests that it implicitly embeds local image semantics and concepts in its intermediate feature maps \cite{zhou2022extract}. 
In this work, we address the challenge of extracting this latent affordance grounding knowledge in a zero-shot manner, \ie without fine-tuning on datasets that explicitly focus on affordance localization task.

To this end, we start from a frozen CLIP model and we introduce a lightweight Feature Pyramid Network (FPN) \cite{Lin_2017_CVPR}, which gradually refines CLIP global descriptor with fine-grained spatial information from early layers of the visual encoder. 
To avoid introducing task-specific biases, we propose to train the FPN on the proxy task of referring image segmentation \cite{wang2022cris, etris,yang2022lavt,zhu2022seqtr}, which provides binary masks of objects, referred by a textual prompt. 
Training on fine-grained segmentation masks exclusively for \emph{objects}, our approach distills CLIP global understanding into pixel-level embeddings without direct action-affordance associations.

Our results demonstrate that our FPN enables the extraction of latent knowledge embedded in CLIP for zero-shot affordance grounding. 
We achieve competitive results w.r.t. existing supervised or weakly-supervised methods, with the additional benefits that: i) we don't need any sort of supervision on actions-objects pairs; ii) we are not bounded to a fixed set of actions and our method can work with open-vocabulary prompts; iii) our method introduces a very limited number of learnable parameters w.r.t. existing solutions. Summarizing, our contributions are the following:
\begin{itemize}
    \item We demonstrate the feasibility to solve affordance segmentation without explicit (weakly-)supervised training;
    \item We showcase how large pre-trained Vision-Language models can naturally handle any action prompts;
    \item To adapt CLIP global descriptors to a dense task without finetuning, we propose a lightweight, low-overhead Feature Pyramid Network to extract multiscale, spatial features while retaining language-aligned embeddings.
\end{itemize}

%% file: sec/2_related.tex
\section{Related works}
\paragraph{Affordance Grounding}
The task of affordance grounding has gained increasing attention in computer vision, seeking to identify image regions that suggest potential interactions between humans and objects. Several works \cite{myers2015affordance,koppula2013learning, do2018affordancenet, chuang2018learning, qian2024affordancellm} have proposed to tackle the problem through supervised approaches, learning to identify relationship maps between local object regions and their associated affordances. However, more recently there has been a consistent effort in searching alternative strategies to mitigate the challenges of collecting costly and extensive annotations.
\cite{sawatzky2017weakly} introduced an innovative weakly supervised approach for affordance detection. By solving an Expectation-Maximization problem \cite{dempster1977maximum}, their methodology relies on a sparse set of key points for weakly supervised affordance detection. \cite{nagarajan2019grounded}, instead, proposes to use affordance labels only, to extract the interactions directly from videos. Notably, \cite{ag_from_exocentric_imgs} annotates the first large-scale affordance dataset - AGD20K, with affordance/object categories and part-level annotations, serving as a benchmark for evaluating the efficacy of different methodologies. Existing weakly supervised object localization and affordance grounding methods \cite{li2023locate, ag_from_exocentric_imgs, ag_from_exocentric_imgs+} are mainly based on class activation mapping  \cite{zhou2016learning} (CAM). 
Unlike traditional methods, our solution avoids the requirement for task-specific, weakly-supervised data by utilizing the knowledge transferred to vision-language models during large-scale pretraining. 

\paragraph{Dense prediction from Vision Language Models}
The shared visual-language embedding space learned from image-text pairs has enhanced open-world detection \cite{ma2024codet,minderer2022simple,liu2023grounding} and segmentation tasks \cite{li2022language,luddecke2021image,zhou2022extract, wang2022cris,rao2022denseclip}. LSeg \cite{li2022language} uses an image encoder trained on labeled segmentation data, which generates pixel-wise embeddings that align with the CLIP text embedding of the corresponding segmentation label. Fine-tuning methods like CRIS \cite{wang2022cris}, CLIPSeg \cite{luddecke2021image} and DenseCLIP \cite{rao2022denseclip} utilize an image decoder to create relevancy maps guided by CLIP text embeddings and the CLIP image encoder. However, the small datasets typically used for fine-tuning often limit the model's broader language understanding. MaskCLIP \cite{zhou2022extract} extracts dense patch-level features from CLIP's image encoder without breaking the visual-language associations. Analogously, our approach directly leverages the multi-modal representation learned by CLIP, without any finetuning of its original parameters.

\label{sec:related}

%% file: sec/3_method.tex
\begin{figure*}[t]
    \centering
    \includegraphics[width=0.95\linewidth]{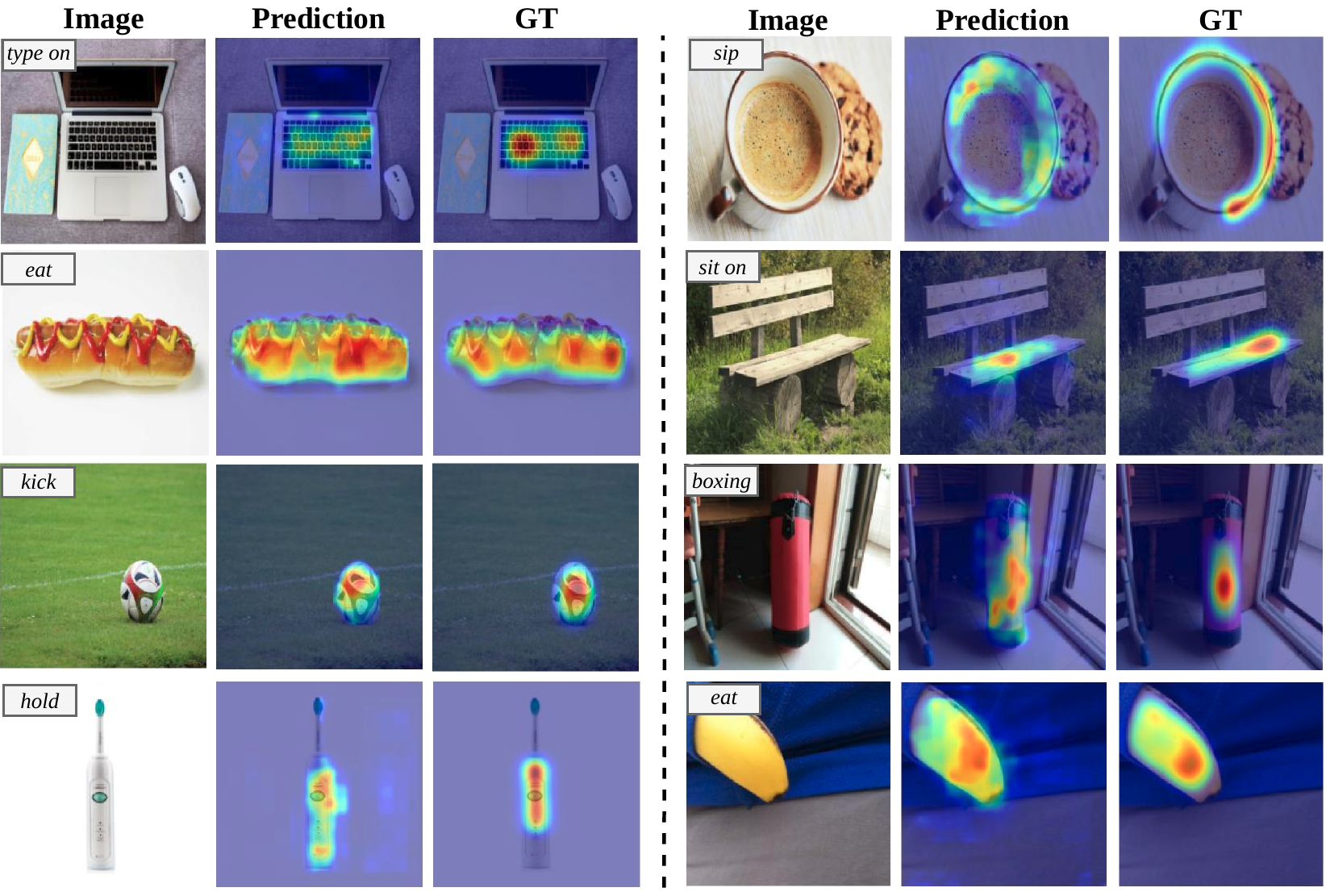}
    \caption{\textbf{Qualitative results}. Given an image and action, we show our model's prediction and the corresponding Ground Truth. } 
    \label{fig:qualitative_good}
\end{figure*}

\section{Method}
This research investigates the potential of CLIP, a powerful pre-trained multimodal model, to identify affordances of objects in an image (\ie affordance grounding). Our framework leverages CLIP pre-trained image-language alignment, refining its output to obtain fine-grained spatial information for accurate localization.

We build on a frozen CLIP model to extract visual and language features, preserving its rich understanding of the relationship between images and text. However, as CLIP processes the image through a deep neural network, the output visual vector loses precise spatial information about where objects are located in relation to each other. Instead, intermediate feature maps extracted by the visual encoder retain both spatial information and local image semantics \cite{zhou2022extract}.  To recover this spatial information, we introduce a Feature Pyramid Network (FPN) \cite{Lin_2017_CVPR} that operates on the CLIP visual encoder at different depths. This FPN gradually integrates spatial details back into the global visual vector, allowing the model to recover crucial object localization information.
Finally, inspired by CLIP original training, we introduce a contrastive learning objective to transfer CLIP  image-level reasoning capabilities at the pixel level. The overall architecture is shown in \cref{fig:architecture}. 
\label{sec:method}

\subsection{Feature extraction}
As feature extractor, we rely on the pre-trained backbone of CLIP to extract semantically aligned visual and textual representation for each input image and corresponding affordance expression.

\paragraph{Image encoder}
Given an input image $I  \in  \mathbb{R}^{H \times W \times 3}$, we extract visual features from an image encoder. Specifically, we employ the frozen ResNet-101 \cite{he2016deep} of CLIP \cite{radford2021learning} to obtain $F_S \in \mathbb{R}^{C}$, where $C$ is the CLIP output dimension. This vector represents a compressed encoding of the image visual content. Additionally, we also consider the hierarchical feature volumes $F_1 \in \mathbb{R}^{\frac{H}{8} \times \frac{W}{8} \times C_1}$, $F_{2} \in \mathbb{R}^{\frac{H}{16} \times \frac{W}{16} \times C_2}$ and $F_{3} \in \mathbb{R}^{\frac{H}{32} \times \frac{W}{32} \times C_3}$, where $C_i$ represents the channel dimension at stage $i$ and $H$ and $W$ are the height and the width, respectively. These features progressively encode higher-level abstractions of the image content. 

\paragraph{Text encoder}
Given the textual query $t$, we extract the tokenized expression $T \in \mathbb{R}^L$, with $L$ being the length of the expression. Note that the tokenization is obtained though lower-cased byte pair encoding (BPE) with 49152 vocabulary size and that the sequence is augmented by adding a global sentence representation token \texttt{[CLS]} and the end of sequence token \texttt{[EOS]}. A Transformer \cite{attention} modified by \cite{radford2021learning} processes $T$ to extract the linguistic features $F_T^i \in \mathbb{R}^{L \times C}$, where $C$ is the number of channels. The activation of the global contextual token \texttt{[CLS]} is further processed to generate the global textual representation $F_Q \in \mathbb{R}^{C}$.

\subsection{Recovering spatial details}
The output vector $F_S$ of CLIP visual encoder captures the global context of the image but lacks the fine-grained details required for highlighting specific objects or regions. 
Pixel-level information is essential for the system to accurately identify contact points, to determine the relative positions of objects, and to analyze their orientations. 

To this end, we introduce a lightweight Feature Pyramid Network (FPN) that enriches CLIP output vector with detailed spatial information.  
Given the hierarchical nature of feature maps, where higher-context information is encoded progressively with lower spatial resolution, we propose to utilize the visual features extracted from the frozen image encoder ($F_1, F_2, F_3$) to augment the CLIP output vector $F_S$. Our approach involves an incremental fusion of lower-resolution features with higher-resolution ones, beginning with $F_S$.

To ensure compatibility across feature dimensions, both the global vector and the hierarchical visual features are first projected into a common representational space $C'$. Specifically, given $F_S$ and the features $F_i$, we compute the projected features:

\begin{alignat}{2}
& F'_S = \texttt{Conv}_{3\times 3}(F_S) \\
& F'_i = \texttt{Conv} _{3\times 3}(F_i), && \ \ i=1,2,3
\end{alignat}
where $\texttt{Conv}_{3\times 3}$ denotes a convolution operation with kernel size 3, followed by a Batch Normalization and ReLU activation.

At this stage, we can directly execute the feature fusion operation. The global vector, $F'_S$, is initially fused with the lowest-resolution feature map, $F'_3$. The fusion process then continues by progressively integrating higher-resolution feature maps, $F'_2$ and $F'_1$.
Formally, we compute the features $F^O_i$ as:
\begin{alignat}{2}
& F^O_1 = F'_1 + \texttt{Up}(F'_S) \\
& F^O_i = F'_i + \texttt{Up}(F^O_{i-1}), && \ \ i=2,3
\end{alignat}
where $\texttt{Up}$ is a parameter-free upsampling operation that increases the resolution of $F^O_{i-1}$ and $F'_S$. 

To obtain the final visual feature, we project the $C'$-dimensional feature representation into the CLIP original feature space $C$. Formally, given the feature $F^O_3$, we compute the output feature $F^O \in \mathbb{R}^{\frac{H}{8} \times \frac{W}{8} \times C}$ as follows:

\begin{alignat}{2}
& F^O = \texttt{Conv}_{1\times 1}(F^O_3)
\end{alignat}
where $\texttt{Conv}_{1\times 1}$ denotes a convolution operation with kernel size 1, followed by a Batch Normalization and ReLU activation.

\subsection{Affordance Head}
By keeping CLIP frozen, the visual-text alignment is preserved. While CLIP condenses  the visual content into a single embedding aligned with a holistic description of the image, through our FPN we expand this representation to a higher resolution, in which individual pixels are semantically aligned with text.
Hence, given the resulting visual features and a textual embedding, the corresponding activation map can be computed with a simple matrix multiplication.

Formally, the activation map $Y_{pred} \in \mathcal{R}^{H \times W}$ is obtained via matrix multiplication between the output of the text encoder ($F_Q$) and the output of the FPN ($F^O$):
\begin{equation}
   Y_{pred} = F_Q \cdot (F^O)^T,
   \label{eq:final_pred}
\end{equation}
where $T$ denotes the transpose operation.

\subsection{Pixel-Text Contrastive Training}
In its pre-training, CLIP employs a contrastive loss to  learn a semantically rich joint representation space for images and their corresponding textual descriptions. The idea is to minimize the distance between the correct image-text associations (\ie their embeddings are pushed closer in the shared space), while maximizing the distance of negative pairs in a batch of images.  
On the other hand, dense predictions tasks such as ours require pixel-level information to delineate the object referred by the query. 

Our FPN is tasked with extracting this information from CLIP intermediate features.
To do so, we apply the same concept of CLIP pre-training, and adopt a contrastive objective on our spatially augmented visual features, to distill CLIP global representation into pixel-level embeddings.
Thus, we use a pixel-text contrastive loss \cite{wang2022cris, etris}, to force the FPN to structure the resulting visual features in such a way that pixels referred by the query are precisely localizable (see \cref{fig:architecture}).

Formally:
\begin{equation}
    L_{con}^{i}=
    \begin{cases}
        -\log \sigma (Y_{pred}^i) & i \in \mathcal{P},\\
        -\log (1 - \sigma (Y_{pred}^i)), & i \in \mathcal{N},\\
    \end{cases}
\end{equation}
\vspace{-2.0mm}
\begin{equation}
    L_{con} = \frac{1}{\left\lvert \mathcal{P} \cup \mathcal{N} \right\rvert} \sum_{i \in \mathcal{P} \cup \mathcal{N}} L_{con}^{i},
    \label{eq10}
\end{equation}
where $\mathcal{P}$ and $\mathcal{N}$ denote the class of ``1'' and ``0'' in the ground truth, $\left\lvert \mathcal{P} \cup \mathcal{N} \right\rvert$ is the cardinality, $\sigma$ is the sigmoid function.

%% file: sec/4_experiments.tex
\input{sec/tables/results_comparison}

\section{Experiments}
\subsection{Datasets}
Following \cite{li2023locate}, we evaluate AffordanceCLIP on AGD20K. To assess CLIP's zero-shot performance in affordance grounding, we strictly avoid using any form of supervision (fully or weakly) derived from this dataset. While CLIP encoders are kept frozen, the Feature Pyramid Network (FPN) requires training to bridge CLIP's image-level reasoning to pixel-level predictions. 
For this reason, we train the FPN exclusively on the RefCOCO/+/g dataset \cite{yu2016modeling, mao2016generation, nagaraja2016modeling}, a popular benchmark for referring image segmentation. RefCOCO/+/g focuses on segmenting objects, described in natural language, rather than affordances. Note that the provided GT are binary  masks, whereas in the downstream task the objective is to obtain continuous activation maps highlighting affordance regions.

\textbf{AGD20K} dataset \cite{ag_from_exocentric_imgs} provides a collection of 20,061 images captured from a third-person perspective (exocentric) and 3,755 images from a first-person perspective (egocentric). These images are annotated with labels for 36 different affordances, which represent the potential objects interactions. 
The AGD20K dataset is designed to evaluate model performance under two settings: Seen and Unseen. In the Seen setting, the categories of objects in the training and test sets are identical. Conversely, the Unseen setting contains novel object categories during testing. Notably, this distinction only applies to methods from the weakly or fully supervised category. In our work, we do not use the any supervision from the affordance dataset and therefore, both splits represent unseen object categories for our model. 
To reflect this, we use a revised nomenclature: \textbf{Test A} corresponds to the original \emph{Seen} setting (1675 images), while \textbf{Test B} corresponds to the original \emph{Unseen} setting (540 images).

\textbf{RefCOCO}, \textbf{RefCOCO+}, and \textbf{RefCOCOg} \cite{yu2016modeling, mao2016generation, nagaraja2016modeling} datasets are widely used benchmarks for evaluating object reference understanding in images. RefCOCO comprises 142,209 short (3.6 words on average) textual descriptions for 50,000 objects in 19,994 images. 
RefCOCO+ introduces a greater challenge with 141,564 descriptions focused purely on appearance-based referencing, deliberately excluding location words. 
RefCOCOg expands the scope with 104,560 longer (8.4 words average) and more complex referring expressions, derived using crowdsourcing through Amazon Mechanical Turk. These expressions reference 54,822 objects across 26,711 images. 

\subsection{Evaluation metrics}
Following \cite{li2023locate, ag_from_exocentric_imgs, ag_from_exocentric_imgs+, qian2024affordancellm}, we evaluate our model in terms of Kullback-Leibler Divergence, Similarity and Normalized Scanpath Saliency.  
\paragraph{Kullback-Leibler Divergence (KLD)} metric quantifies the discrepancy between the predicted affordance distribution ($M$) and the ground truth distribution ($M'$).

\begin{equation}
   \mathrm{KLD}\left ( M,M' \right )=\sum_{i}M'_{i}\log\left ( \epsilon + \frac{M'_{i}}{\epsilon+M_{i}} \right ), \label{eq:no20}
\end{equation}

\paragraph{Similiary (SIM)} measures the intersection between the predicted affordance map ($M$) and the ground truth ($M'$).

\begin{equation}
   \mathrm{SIM}\left ( M, M' \right )=\sum_{i}\min\left ( M_{i},M'_{i}\right ),\\
\end{equation}
where  $\sum_{i}M_{i}=\sum_{i}M'_{i}=1$.

\paragraph{Normalized Scanpath Saliency (NSS)} measures the correspondence between the prediction map ($M$) and the ground truth ($M'$).

\begin{equation}
   \mathrm{NSS}\left ( M,M' \right )=\frac{1}{N}\sum_{i}\hat{M}\times M'_{i}, \label{eq:no22}
\end{equation}
where $N=\sum_{i}M'_{i}$, $\hat{M}=\frac{M-\mu\left ( M \right )}{\sigma\left ( M \right )}$. $\mu\left ( M \right )$ and $\sigma\left ( M \right )$ are the mean and standard deviation, respectively.

\subsection{Implementation Details}
We initialize the text and image encoder with CLIP, adopting ResNet-101 as visual encoder. The FPN is trained for 1 epoch with a batch size of 32 on a combination of RefCOCO, RefCOCO and RefCOCO+ images.  Input images are resized to $416 \times 416$, following \cite{wang2022cris,etris}. We use Adam optimizer with a learning rate of $\lambda = 0.0001$.

%% file: sec/tables/results_comparison.tex
\begin{table*}[!htb]
    \centering
    \begin{tabular}{clccc@{\hskip 0.3in}ccc}
    \toprule
    \multicolumn{2}{c}{\multirow{2.5}{*}{\textbf{State-of-the-Art from Relevant Tasks}}} &  \multicolumn{3}{c@{\hskip 0.3in}}{\textbf{Test A (\textit{Seen})}} & \multicolumn{3}{c@{\hskip 0.15in}}{\textbf{Test B (\textit{Unseen})}} \\ \cmidrule(lr@{0.3in}){3-5} \cmidrule(r){6-8}
                                                                                                  &                         & KLD$\downarrow$    & SIM$\uparrow$    & NSS$\uparrow$  & KLD$\downarrow$     & SIM$\uparrow$     & NSS$\uparrow$    \\ \midrule
    \rowcolor{mylightgray}
    \multicolumn{8}{l}{Fully Supervised Affordance Grounding} \\ \midrule
    & AffordanceLLM \cite{qian2024affordancellm} & - & - & - & 1.463  & 0.377  & 1.070   \\ \midrule
    \rowcolor{mylightgray}
    \multicolumn{8}{l}{Weakly Supervised Object Localization*} \\ \midrule
    & EIL \cite{EIL} & 1.931  & 0.285  & 0.522 & 2.167   & 0.227   & 0.330  \\
    & SPA \cite{SPA} & 5.528  & 0.221  & 0.357 & 7.425   & 0.169   & 0.262  \\
    & TS-CAM \cite{TS-CAM}  & 1.842  & 0.260  & 0.336 & 2.104   & 0.201   & 0.151  \\ \midrule
    \rowcolor{mylightgray}
    \multicolumn{8}{l}{Weakly Supervised Affordance Grounding} \\ \midrule
    & Hotspots \cite{grounded} & 1.773  & 0.278  & 0.615 & 1.994   & 0.237   & 0.577  \\
    & Cross-view-AG \cite{ag_from_exocentric_imgs} & 1.538  & 0.334  & 0.927 & 1.787   & 0.285   & 0.829  \\
    & Cross-view-AG+ \cite{ag_from_exocentric_imgs+} & 1.489  & 0.342  & 0.981 & 1.765  & 0.279   & 0.882  \\
    & Locate \cite{li2023locate} & 1.226 & 0.401  & 1.177 & 1.405   & 0.372   & 1.157 \\ \midrule
    \rowcolor{mylightgray}
    \multicolumn{8}{l}{Zero-shot Affordance Grounding} \\ \midrule
    & \textbf{AffordanceCLIP}  &   1.628 & 0.335 & 0.791 & 1.812 & 0.301 & 0.760 \\
    
    \bottomrule
    \end{tabular}
    \caption{Comparison to state-of-the-arts methods on AGD20K dataset. Following LOCATE \cite{li2023locate}, we include state-of-the-art methods from a relevant task - weakly supervised object localization. Results of * are taken from \cite{ag_from_exocentric_imgs}. ($\uparrow$/$\downarrow$ means higher/lower is better).}
    \label{tab:com_sota}
    \end{table*}

%% file: sec/5_results.tex
\section{Results}
\label{sec:results}
\input{sec/tables/params}

\subsection{Comparison with State-of-the-art}
In order to provide a comprehensive benchmark, we consider state-of-the-art methods on the affordance grounding task under varying supervision levels: fully supervised and weakly supervised. Additionally, following LOCATE \cite{li2023locate}, we include results from state-of-the-art methods from the related task of weakly supervised object localization. Results are presented in \cref{tab:com_sota}.

Our results showcase the strong generalization capabilities of AffordanceCLIP on the affordance grounding task. This suggests that even though CLIP itself was not explicitly trained for this task, it has implicitly captured relevant visual features and their relationships to concepts, including actions and interactions. 
Remarkably, AffordanceCLIP outperforms Weakly Supervised Object Localization approaches, on both the test splits, and is competitive with the Affordance Grounding methods that leverage weakly supervised data.
Furthermore, \cref{tab:param} emphasizes the efficiency of our model. By training only a lightweight Feature Pyramid Network on top of CLIP, we significantly reduce the number of trainable parameters compared to competing approaches. Notably, the FPN, with only 2.71 M parameters, effectively bridges the gap between CLIP's global image understanding and the pixel-level precision required for affordance grounding.

\begin{figure*}[t]
    \centering
    \includegraphics[width=0.7\linewidth]{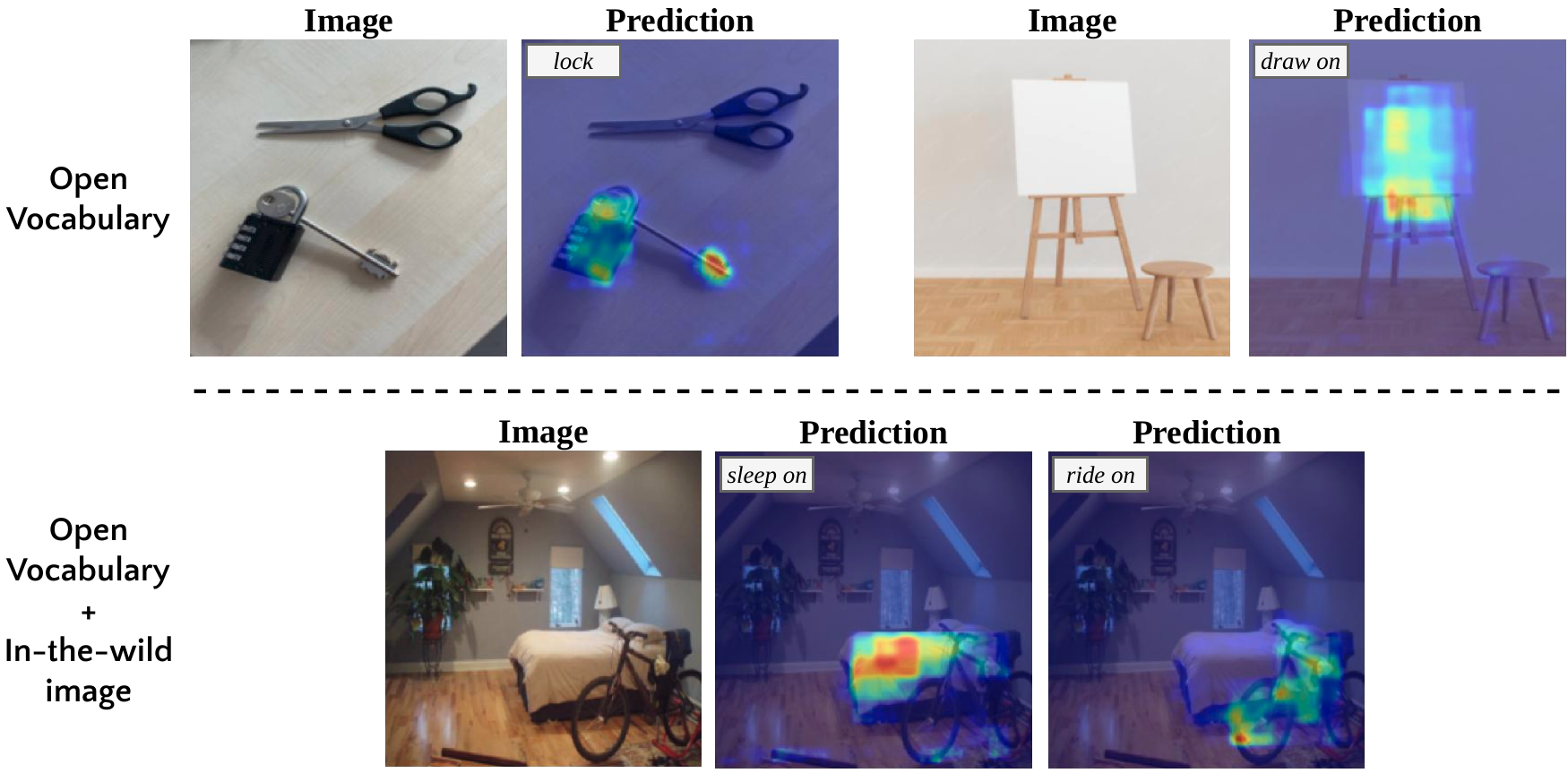}
    \caption{\textbf{Open Vocabulary capabilities.} \textit{Top}: AffordanceCLIP is queried with actions outside the 36 of AGD20K dataset. \textit{Bottom}: AffordanceCLIP is tested \textit{in the wild}, on a challenging image from everyday settings.} 
    \label{fig:qualitative_ov}
\end{figure*}

\input{sec/tables/ablations}

\subsection{Ablation study}
To analyze the contribution of different levels of spatial detail from CLIP's intermediate ResNet-101 \cite{resnet} features, we conducted an ablation study.  \cref{tab:ablation} summarizes the results, where we progressively integrate higher resolution feature maps into the FPN. Results demonstrate a consistent performance improvement as we integrate additional, more spatially detailed features. This suggests that each feature map provides valuable complementary information, enhancing the model's ability to perform accurate localization of affordance regions. This experimental evidence confirms the value of latent knowledge encoded within CLIP's intermediate representations.

\subsection{Qualitative results}
In \cref{fig:qualitative_good}, we present qualitative results that highlight the remarkable capabilities of AffordanceCLIP.  These results demonstrate the model's capabilities in two key areas. First, it accurately localizes the target object within the image, successfully differentiating it from other visually similar or contextually related objects. Second, AffordanceCLIP precisely identifies the specific region within the object where the queried affordance can be performed. Consider the query \textit{type on}: AffordanceCLIP is first of all able to discern between multiple objects in the image (the mouse, the laptop, the notebook) to identify the object associated with the action; then, it disambiguates within the regions of the object (the display, the keyboard, the touchpad) to identify the part with which we perform the action.

\subsection{Open-Vocabulary capabilities}
In \cref{fig:qualitative_ov} we qualitatively demonstrate the open-vocabulary capabilities of AffordanceCLIP, by testing our model with new actions outside those present in the AGD20K dataset. Results show that our adaptation of CLIP to dense predictions has not compromised its knowledge of open-world concepts.
For example, it can be queried with \textit{lock} or \textit{draw on} without requiring additional human-object interaction images or model finetuning for these actions. 
Due to its pre-training on complex scenes, CLIP demonstrates robust performance on in-the-wild images from the Internet featuring challenging, everyday settings. For example, in \cref{fig:qualitative_ov} (bottom), AffordanceCLIP successfully interprets a complex bedroom scene containing diverse objects and an unusual configuration (a bicycle near the bed). AffordanceCLIP is able to identify the bed when queried with \textit{sleep on}, but also the bike if prompted with \textit{ride on}.

\section{Limitations}
\begin{figure}[t]
    \centering
    \includegraphics[width=0.9\linewidth]{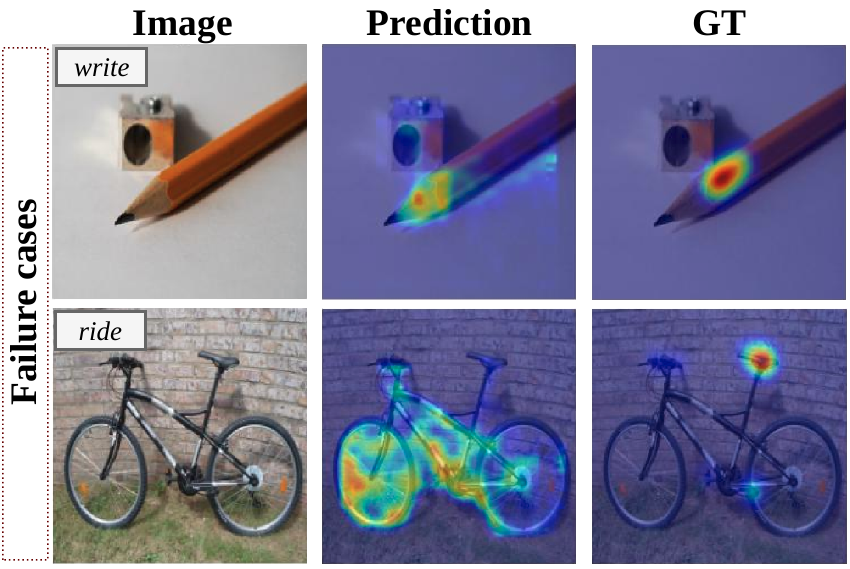}
    \caption{Examples of failure cases. } 
    \label{fig:qualitative_bad}
    \vspace{-0.3cm}
\end{figure}

Despite AffordanceCLIP's impressive zero-shot affordance grounding abilities, it does have limitations. 
\cref{fig:qualitative_bad} highlights some interesting scenarios where it fails. In one case, when asked to identify a region to \textit{write on}, the model focuses on the pencil tip rather than the part of the pencil we grasp with our hand. This suggests that  CLIP has strongly associated the concept of writing with the tool used for the action, rather than the part of the object directly manipulated.  Another interesting failure occurs when prompted with \textit{ride}. AffordanceCLIP correctly locates the bike but excludes the bike seat.
This weaker association between the seat and the concept of riding may be due to the fact that in many images used to train CLIP, the seat is often occluded by the rider.

\section{Conclusions and future works}
In this work, we explored an alternative approach to affordance grounding. We move away from traditional weakly-supervised learning methods and instead leverage the implicit knowledge within visual language models (like CLIP) to identify object activation regions based on action prompts. We believe that our work paves the way for future research towards  open-vocabulary affordance grounding. These promising results highlight the potential of exploring more advanced Vision-Language models, like LLaVA \cite{liu2024visual}, Flamingo \cite{alayrac2022flamingo} and GPT-4V \cite{yang2023set}. In particular, these models are able to answer questions which require a deeper understanding of objects and their relationship to abstract concepts. 
This level of reasoning is essential in affordance grounding when dealing with complex images and queries, which demand taking into account object properties beyond mere geometric shape (\eg material, inertial parameters) to associate them to functionalities.

%% file: sec/tables/params.tex
\begin{table}[!t]
    \centering
    \begin{tabular}{lcc}
    \toprule
    Methods         & Params (M) \\ \midrule
    EIL \cite{EIL}            & 42.41   \\
    SPA \cite{SPA}            & 69.28   \\
    TS-CAM \cite{TS-CAM}         & 85.86 \\ \midrule
    Hotspots \cite{grounded}       & 132.64    \\
    Cross-view-AG \cite{ag_from_exocentric_imgs}  & 120.03   \\
    Cross-view-AG+ \cite{ag_from_exocentric_imgs+} & 82.27     \\
    Locate \cite{li2023locate} & 6.50   \\
    \textbf{AffordanceCLIP} & \textbf{2.71}
    \\\bottomrule
    \end{tabular}
    \caption{Comparison of learnable parameters.}
    \label{tab:param}
    \vspace{-0.2cm}
\end{table}

%% file: sec/tables/ablations.tex
\begin{table}[ht]
    \begin{center}
    \begin{adjustbox}{width=\linewidth}
    \begin{tabular}{lcccccc}
    \toprule
    \multirow{2.5}{*}{Method} & \multicolumn{3}{c}{Seen} & \multicolumn{3}{c}{Unseen} \\ 
    \cmidrule(lr){2-4} \cmidrule(l){5-7} 
        & KLD$\downarrow$    & SIM$\uparrow$    & NSS$\uparrow$    & KLD$\downarrow$     & SIM$\uparrow$     & NSS$\uparrow$    \\ 
    \midrule
    \{$F_1^O$\}   &		1.917 &	0.322 &	0.665 & 2.171	& 0.278 &	0.586    \\ 
    \{$F_1^O$, $F_2^O$\}  	&	1.892 &	0.329 &	0.726 &   2.038 &	0.297	& 0.696     \\
    \{$F_1^O$, $F_2^O$, $F_3^O$\}   &   \textbf{1.628} & \textbf{0.335} & \textbf{0.791} & \textbf{1.812} & \textbf{0.301} & \textbf{0.760}      \\ \bottomrule
    \end{tabular}
    \end{adjustbox}
    \end{center}
    \vspace{-0.4cm}
    \caption{Contribution of different levels of spatial detail from CLIP’s intermediate ResNet-101 features.}
    \label{tab:ablation}
    \vspace{-0.2cm}

\end{table}

%% file: sec/6_ack.tex
\newcommand{\myparagraph}[1]{\vspace{4pt}\noindent\textbf{#1}}

\small{
\myparagraph{Acknowledgements}
This study was carried out within the Sustainable Mobility Center (CNMS) and received funding from the European Union Next Generation EU (Piano Nazionale di Ripresa e Resilienza (PNRR), Missione 4 Componente 2 Investimento 1.4 "Potenziamento strutture di ricerca e creazione di "campioni nazionali di R\&S" su alcune Key Enabling Technologies") with grant agreement no. CN\_00000023. 
We also acknowledge  FAIR - Future Artificial Intelligence Research which received funding from the European Union Next-GenerationEU (PIANO NAZIONALE DI RIPRESA E RESILIENZA (PNRR) – MISSIONE 4 COMPONENTE 2, INVESTIMENTO 1.3 – D.D. 1555 11/10/2022, PE00000013). 
This manuscript reflects only the authors’ views and opinions, neither the European Union nor the European Commission can be considered responsible for them.}